\documentclass{article}
\usepackage{arxiv}
\usepackage{times}
\usepackage{url}
\usepackage{latexsym}
\usepackage{booktabs} 
\usepackage{multirow}
\usepackage{graphicx} 
\usepackage{bbm} 
\usepackage{adjustbox} 
\usepackage{amsmath} 
\usepackage{xcolor}
\usepackage{natbib}
\definecolor{darkblue}{rgb}{0, 0, 0.5}
\usepackage{hyperref}
\hypersetup{colorlinks=true,citecolor=darkblue, linkcolor=darkblue, urlcolor=darkblue}

\title{Pre-trained language models as knowledge bases for Automotive Complaint Analysis}

\author{
  Vanessa Deborah Viellieber \\
  Data Science and Artificial Intelligence \\
  MHP - A Porsche Company \\
  Ludwigsburg, Germany \\
  \texttt{vanessa.viellieber@mhp.com}\hspace{.5cm} \\\And
  Matthias~Aßenmacher \\
  Department of Statistics\\
  Ludwig-Maximilians-Universität \\
  Munich, Germany \\
  \texttt{matthias@stat.uni-muenchen.de} \\
}

\begin{document}

\maketitle

\begin{abstract}
Recently it has been shown that large pre-trained language models like BERT \citep{devlin2018bert} are able to store commonsense factual knowledge captured in its pre-training corpus \citep{petroni2019language}. In our work we further evaluate this ability with respect to an application from industry creating a set of probes specifically designed to reveal technical quality issues captured as described incidents out of unstructured customer feedback in the automotive industry.
After probing the out-of-the-box versions of the pre-trained models with fill-in-the-mask tasks we dynamically provide it with more knowledge via continual pre-training on the Office of Defects Investigation (ODI) Complaints data set. In our experiments the models exhibit performance regarding queries on domain-specific topics compared to when queried on factual knowledge itself, as \citet{petroni2019language} have done. For most of the evaluated architectures the correct token is predicted with a $Precision@1$ ($P@1$) of above 60\%, while for $P@5$ and $P@10$ even values of well above 80\% and up to 90\% respectively are reached. These results show the potential of using language models as a knowledge base for structured analysis of customer feedback.
\end{abstract}

\section{Introduction}\label{sec:intro}

Recently researchers developed some interest in the knowledge stored in the large pre-trained models. \citet{petroni2019language} investigated BERT \citep{devlin2018bert} and other architectures with respect to their ability of storing commonsense factual knowledge. As the stored knowledge depends heavily on the pre-training corpus, we are curious about whether one can "teach" these kinds of models further knowledge by exposing them to texts from specific domains, like customer complaints in the automotive industry.

Especially for product-driven organizations as car manufacturers, customer feedback provides a precious source of information for product improvements, e.g. in terms of potential security risks identified and mentioned by customers. However, the structured use of this data is an open problem in industry, despite numerous investigations with advanced NLP methods \citep{choe2013semiautomated,lee2015ontology,akella2017gain,liang2017mining,joung2019customer}. Handling this fuzzy data and satisfying the demand for detailed information extraction in an intelligent manner remains challenging.

The recent developments in NLP lead us to the idea of evaluating the ability of pre-trained language models to act as a domain-specific knowledge base. We investigate if a language model, further pre-trained on customer feedback, is able to store customer opinions about products, features, and services as knowledge in model parameters.

\section{Related work}\label{sec:rel}

Besides the challenges of vast customer complaints with free flow writing, different languages, abbreviations, misspellings, domain-specific entities, what makes the analysis of customer complaints so heavily complex is that customer issues occur in so many different forms and combinations. With supervised learning methods the limits are that customer feedback analysis only works along a certain number of categories \citep{liang2017mining,akella2017gain}. With these methods it is impossible to automatically identify newly emerging, possibly security-relevant, risks. The lack of labeled training data as well as imbalanced data is a hurdle in the development of NLP models in industrial customer feedback analysis \citep{choe2013semiautomated,akella2017gain}. Furthermore, it is relevant for product manufacturers to be able to identify implied connections in customer feedback. \citet{lee2015ontology} used Web Ontology Language (OWL) to express semantic relations in the customer complaint analysis. But schema engineering strongly needs human supervision and is very time-consuming \citep{lee2015ontology,wang2016novel}. Therefore this depicts an expensive process (for industrial application) that some product manufacturers shy away from and thus leave a great deal of information unused.

As recently observed by \citet{petroni2019language} and \citet{zhang2019ernie}, language models can store implicit knowledge after pre-training and thus act as a kind of knowledge base, which could be a solution for the predominant challenges in customer complaint analysis. Therefore we use these findings to create domain specific probes for automotive industry in the style of \citet{petroni2019language}. They used a general corpus of facts representing knowledge statements like "\textit{iPod Touch is produced by Apple}". After converting these facts into cloze statements they query the language model asking it to fill in a masked token, which was always the \textit{object} of a fact triple (\textit{subject - predicate - object}). For the evaluation of the models they determined how high the ground truth token was ranked against every other word in a fixed candidate vocabulary. 

\section{Materials and Methods}\label{sec:methods}

We evaluate a selection of pre-trained language models for the English language which recently achieved state-of-the-art results on frequently used benchmarks, for which we use the stable implementations via the unified API of the \texttt{transformers}\footnote{\url{https://github.com/huggingface/transformers}} module \citep{wolf2019huggingface}. As we want to infer the effect of continual pre-training on a domain-specific corpus, we use the architectures with the respective heads (e.g. \texttt{BertForMaskedLM}\footnote{\url{https://huggingface.co/transformers/model\_doc/bert.html\#bertformaskedlm}}). 

As in-domain corpus for the continual pre-training we use a collection of roughly 500.000 publicly available e-mails containing customer complaints of various vehicle makes. Since this corpus should inherit many details which indicate deeper knowledge of the original equipment manufacturers (OEMs) products, we suspect that continual pre-training transfers this knowledge into the model's weights.

The evaluation set of probes is handcrafted from a held-out set from the same corpus which we utilize for continual pre-training. We created a dictionary containing technical terms of automotive industry and subsequently filtered the corpus using this dictionary in order to obtain suitable sentences where model parts could be masked for probing the model.
 
\subsection{Pre-trained language models}\label{ssec:lms}

BERT \citep{devlin2018bert} is a bidirectionally contextual transformer encoder model, which is available in a \texttt{base} ($\sim$ 110M parameters) and a \texttt{large} ($\sim$ 340M parameters) variant. The \texttt{base, uncased} variant will be used as baseline model. The \texttt{cased} variants for both model sizes are not considered since the raw data is only available capital letters and is thus lower-cased during pre-processing. The competing RoBERTa model \citep{liu2019roberta} is an optimized, but architecturally alike, version of BERT pre-trained on a significantly larger corpus. We will also compare BERT to DistilBERT \citep{sanh2019distilbert} and ALBERT \citep{lan2019albert}, two models which employ parameter reduction to the original BERT architecture. While DistilBERT relies on knowledge distillation \citep{hinton2015distilling}, ALBERT primarily makes use of cross-layer parameter sharing techniques \citep{lan2019albert}.

\subsection{Data for continual pre-training}\label{ssec:data}

We use data from the ODI\footnote{\url{https://catalog.data.gov/dataset/nhtsas-office-of-defects-investigation-odi-complaints}} of the National Highway and Traffic Safety Administration (NHTSA). NHTSA-ODI's data set consists, amongst other sources, of vehicle owner's complaints regarding different manufacturers and is used to identify safety issues that warrant investigation and to determine if a safety-related defect trend exists. Other sources are e.g. consumer action groups or insurance companies. In our work, we filtered the data base only for direct customer complaints in order to obtain a data set most alike to customer complaints as they are addressed to OEMs and documented in their CRM system.

The final corpus consists of 502.445 distinct complaints and has a size of 142.776 kilo bytes. We divide the data with a ratio of 90 to 10 into training and held-out set. The whole corpus has a vocabulary of 139.982 distinct words on which we, besides lower-casing, did not perform any other pre-processing steps. Descriptive statistics are displayed in Tab. \ref{tab:len}, we use the respective tokenizers for the models. 

\begin{table}[ht]
    \centering
    \begin{adjustbox}{max width=1\textwidth}
\begin{tabular}{|c||cccccc|}
\toprule
 & avg. length & min & 25\% & 50\% & 75\% & max \\
\midrule \midrule
 \texttt{raw data} & 35.94 & 1 & 30 & 41 & 43 & 71 \\
 \texttt{tokenized (BERT base uncased)} & 45.47 & 1 & 38 & 51 & 54 & 91 \\
  \texttt{tokenized (BERT large uncased)} & 45.55 & 1 & 38 & 51 & 54 & 106 \\
 \texttt{tokenized (RoBERTa base)} & 49.33 & 3 & 42 & 54 & 58 & 124 \\
  \texttt{tokenized (DistilBERT base uncased)} & 45.55 & 1 & 38 & 51 & 54 & 106 \\
 \texttt{tokenized (ALBERT xxlarge)} & 48.70 & 3 & 40 & 54 & 58 & 130 \\
\bottomrule
\end{tabular}
    \end{adjustbox}
    \caption{Descriptives for the raw and the tokenized sequences of the train set. BERT uses a word-piece tokenizer, while ALBERT uses sentence-piece. RoBERTa and DistilBERT rely on a BPE-tokenizer.} 
    \label{tab:len}
\end{table}

\subsection{Set of evaluation probes}\label{ssec:probe}

We extracted the affected components from the variable \texttt{compdesc} (\textit{description of affected components}) in ODI's data set. To stick to the one-token logic of \citet{petroni2019language}, we split up compounds and thus get a dictionary of 364 distinct, mainly technical, terms related to the vehicle. Due to the decomposition of the compounds, it is sometimes the case that individual terms do not necessarily describe a vehicle component unambiguously, but are nevertheless mainly of a technical nature or relevant in automotive customer complaints. An excerpt of the most frequent ten entries can be seen in Tab. \ref{tab:most}.

\begin{table}[ht]
    \centering
    \begin{adjustbox}{max width=1\textwidth}
\begin{tabular}{|c||cccccccccc|}
\toprule
 \texttt{term}      & engine & hydraulic & service & brakes & system & cooling & antilock & air & power & seat \\
\midrule
 \texttt{frequency} & 42903  & 31861     & 30058   & 30044  & 27874  & 17396   & 16391    & 16301 & 16154 & 15874 \\
\bottomrule
\end{tabular}
    \end{adjustbox}
    \caption{Most frequent technical terms to be replaced for creating the probes.} 
    \label{tab:most}
\end{table}

\noindent We identified the sentences which contain the relevant terms and masked one of these terms per sequence to prepare for the masked language modeling task. Tab. \ref{tab:example} shows exemplary evaluation probes.

\begin{table}[ht]
    \centering
    \begin{adjustbox}{max width=1\textwidth}
\begin{tabular}{|c||c|}
\toprule
 \texttt{original} & \textit{gear shift cable failure in auto transmission} \\
 \midrule
 \texttt{masked}   & \textit{\texttt{[CLS]} \texttt{[MASK]} shift cable failure in auto transmission \texttt{[SEP]}} \\
 \texttt{masked}   & \textit{\texttt{[CLS]} gear \texttt{[MASK]} cable failure in auto transmission \texttt{[SEP]}} \\
\bottomrule
\end{tabular}
    \end{adjustbox}
    \caption{Exemplary probe from our test set. One original sentence can lead to multiple probes if it contains more than one affected component. Multi-token components are not replaced by one \texttt{MASK}.} 
    \label{tab:example}
\end{table}

\section{Results}\label{sec:results}

We evaluate the models at regularly spaced intervals defined by the number of in-domain examples seen by a model. Performance values for all five different models are displayed in Tab. \ref{tab:perf}.

\begin{table}[ht]
    \centering
    \begin{adjustbox}{max width=1\textwidth}
\begin{tabular}{|c|c||ccccc|}
\toprule
 & \#in-domain examples & out-of-the-box & 100k & 200k & 300k & 400k \\
\midrule \midrule
 \multirow{2}{*}{\rotatebox[origin=l]{90}{BERT}} 
 & \texttt{base, uncased} & 12.5 (25.8/ 30.2) & 44.5 (71.5/ 78.3) & 49.5 (74.2/ 80.5) & 58.2 (80.5/ 85.6) & 59.0 (81.6/ 86.4) \\
 & \texttt{large, uncased} & 27.1 (57.1/ 65.0) & 49.7 (74.1/ 80.6) & 55.0 (78.3/83.9) & \textbf{64.6} (\textbf{85.2}/ 89.0) & \textbf{63.8} (\textbf{84.8}/ 88.8) \\
 \midrule
 \multirow{4}{*}{\rotatebox[origin=l]{90}{Others}} 
 & RoBERTa (\texttt{base}) & 35.8 (61.8/ 71.1) & 52.2 (79.9/ 87.7) & 51.8 (80.0/ 88.6) & 51.2 (80.2/ 89.1) & 51.5 (80.8/ 89.2) \\
 & DistilBERT (\texttt{base}) & 25.5 (47.8/ 57.4) & 56.5 (78.4/ 84.0) & 58.1 (79.4/ 84.8) & 60.0 (81.1/ 86.1) & 61.1 (82.1/ 86.9) \\
 & ALBERT (\texttt{xxlarge}) & \textbf{43.3 (70.0/ 78.3)} & \textbf{60.4 (83.7/ 89.8)} & \textbf{59.3 (83.7/ 90.7)} & 60.6 (84.4/ \textbf{91.2}) &  59.0 (84.0/ \textbf{91.2}) \\
\bottomrule
\end{tabular}
    \end{adjustbox}
    \caption{Model performances on the test set measured as Precision@k $P@1$ ($P@5$/$P@10$) in percent for all considered architectures. Results are displayed separately for different amounts of additional in-domain examples used for continual pre-training. Best performance per column in bold.} 
    \label{tab:perf}
\end{table}

Overall we can see that across all different models there is only little ability for successfully performing this task when simply used "out-of-the-box", only ALBERT \citep{lan2019albert} already consistently predicts domain relevant terms. E.g. for the second masked sequence from Tab. \ref{tab:example} ALBERT's "out-of-the-box" version predicts the following top five tokens: (1) '\textit{wrench}', (2) '\textit{axle}', (3) '\textit{shifter}', (4) '\textit{shift}' (ground truth), (5) '\textit{switch}' which contain the ground truth on rank four. The other models do not predict solely technical terms in such a consistent way. When pre-trained on our domain-specific corpus in a continual fashion, we can see that performance steadily increases for each of the examined models, partly by a large margin. After having seen 100k domain-specific examples, the values for $P@1$ stagnate for all of the architectures. For RoBERTa and ALBERT this happens rather early (after between 100k and 200k examples), while both BERT variants as well as DistilBERT still show smaller improvements until having seen all 400k examples. When relaxing the performance measure by considering $P@5$ and $P@10$ we observe similar behavior for the BERT variants and for DistilBERT. RoBERTa and ALBERT (contrary to before) are now also showing steady, decreasing improvements until the end of continual pre-training. The superiority of ALBERT, which we already observe for the "out-of-the-box" version, mostly prevails over the course of the evaluation intervals. Only the \texttt{large} variant of BERT is able to compete when having seen many (300k -- 400k) in-domain examples. Concerning the choice of a suitable performance measure, $P@1$ might be a little bit too harsh, since in some cases plausible alternative are ranked above the ground truth. This is a point which requires further evaluation of qualitative nature.

\section{Discussion}\label{sec:disc}

There are several limitations which we have not yet been able to address. First and foremost multi-token words represent a problem, especially in our domain of application. A significant amount of car parts (e.g. \textit{coolant system}) and key issues (e.g. \textit{not working properly}) are not represented by single tokens, but instead have multi-token representations. Following \citet{petroni2019language}, probing \textit{multi-tokens} and \textit{predicates} remains an open challenge. Further extensions of this work will try to address this shortcoming. In addition, we would like to investigate the effects on the model when focal points of different components are described by the same defect patterns, under which conditions the model is then able to differentiate. Moreover we are interested in the time trend within the language models: Can knowledge for a certain temporal period be mapped or queried as such? It is also necessary to investigate the velocity with which new technical issues can be identified in a knowledge base.

\section{Conclusion}\label{sec:conc}

We took an approach already investigated by \citet{petroni2019language} on some tentative probes for commonsense factual knowledge and extended it via continual pre-training on a corpus from a specific domain. Our experiment shows promising results which indicate that pre-trained language models have the ability of representing focal points gathered from customer complaints and are therefore able to represent domain-specific knowledge. Concluding, language models could be an innovative approach to handle the predominant problems in industry to utilize the full potential of unstructured information in customer feedback at economically acceptable expense. According to our industry experience, both the problem and the solution can be transferred from the automotive domain to other product-driven industries.  Nevertheless, there are still some severe limitations to the use of language models as a customer opinion base. These limitations require further investigation and have to be overcome eventually in order to replace scheme-based knowledge bases with these self-supervised approaches.

\clearpage

\bibliography{references}
\bibliographystyle{apalike}

\end{document}